\documentclass[]{llncs}

\usepackage[utf8]{inputenc}
\usepackage[english]{babel}

\usepackage[dvipdfmx]{graphicx}
\usepackage{amsmath}
\usepackage{amssymb}
\usepackage{bm}
\usepackage{listings}
\usepackage{color}
\usepackage{url}
\usepackage{afterpage}
\usepackage{subfigure}
\usepackage{float}
\usepackage{multirow}
\usepackage{bussproofs}
\usepackage{intmacros}
\usepackage{algorithmic}

\usepackage{article}

\begin{document}

\mainmatter

\title{Scalable Parallel Numerical CSP Solver} 
\author{Daisuke Ishii\inst{1}, Kazuki Yoshizoe\inst{2,1} \and Toyotaro Suzumura\inst{3,2}}
\institute{Tokyo Institute of Technology, 
  Tokyo, Japan. \and
  Japan Science and Technology Agency, Japan. \and
  IBM Research, Dublin, Ireland. \\
  \email{\{dsksh,yoshizoe,suzumura\}@acm.org}}

\maketitle

\begin{abstract} 
We present a parallel solver for numerical constraint satisfaction problems (NCSPs) that can scale on a number of cores.
%
Our proposed method 
runs worker solvers on the available cores and
simultaneously the workers cooperate for the search space distribution and balancing.
In the experiments, we attained up to 119-fold speedup using 256 cores of a parallel computer.
\end{abstract}


\section{Introduction}

Numerical constraint satisfaction problems (NCSPs, Section~\ref{s:ncsp})
have been successfully applied to problems described in the domain of reals%
\cite{CCGIJ2014,Ishii2012}.
%
%
Given a NCSP with search space represented as a \emph{box} (i.e., interval vector),
the \emph{branch and prune algorithm} 
efficiently computes a \emph{paving}, a set of boxes that encloses the solution set, yet its exponential computational complexity limits the tractable instances.
Although the solving process 
exhibits a parallelism,
no parallel NCSP solver has been made available to date
because of the difficulty in partitioning the search space equally.

In this research, we parallelize a NCSP solver to scale its solving process on
both shared memory and distributed memory parallel computers (Section~\ref{s:parallelization}).
Our parallel method consists of parallel worker solvers that solve a portion of search space on CPU cores and interact with neighbor workers via message passing for dynamic load balancing. 
We also propose a preprocess that accelerates the initial search space distribution by sending sets of boxes via static routing between the workers.
%
We have implemented the method by extending the Realpaver solver using the X10 language to realize a process-level parallelization over a number of cores.
Section~\ref{s:exp} reports experimental results when our method was deployed on
two hardware environments.

\Subsubsection{Related work.}

There have been several work regarding parallel solving of CSPs with either discrete 
or continuous domains.
Parallel solving of generic CSPs on massive computer clusters and supercomputers has been
explored in \cite{Jaffar2004,BH2006,Xie2010}.
This work focuses on a massive parallel solver for NCSPs
that has a different characteristics compared to generic CSPs.
In the survey~\cite{Gent2011}, existing work is classified into
(i) search-space splitting methods\cite{Schulte2000,Jaffar2004,BH2006,Chu2009,Schubert2009,Xie2010},
(ii) cooperative methods for heterogeneous solver agents (cf. portfolios)\cite{BH2006}, and
(iii) parallelization of constraint propagation.
Our method belongs to the first category.
A few works have used approaches (ii)\cite{Granvilliers2000} 
and (iii)\cite{Goldsztejn2010} for parallelization of NCSP solving.
However, to the best of our knowledge, a massive parallelization method that uses the typical approach (i) has not yet been proposed.

Substantial work regarding the parallelization of the \emph{branch and bound} algorithm
with search-space splitting exists~\cite{Grama2003,Luling1996,Otten2010}.
This approach has also been applied to CSP solvers~\cite{Schulte2000,Chu2009} and 
SAT solvers~\cite{
Schubert2009}.
This work explores an efficient parallel method for solving NCSPs with similar approach to \cite{Grama2003,Luling1996}.

%
%

\section{Numerical Constraint Satisfaction Problems}
\label{s:ncsp}

%
%
%
A \emph{numerical constraint satisfaction problem} (NCSP) is defined as a triple $(v,\vv_0,c)$ that consists of
a vector of \emph{variables} $v = (v_1,\ldots, v_{n})$,
an \emph{initial domain} in the form of a box $\vv_0 \in \IntSet^{n}$ ($\IntSet$ denotes the set of closed real intervals), and
a \emph{constraint} $c(v) \equiv f(v)=0 \land g(v)\geq0$, where $f:\RealSet^{n}\to\RealSet^{e}$ and $g:\RealSet^{n}\to\RealSet^{i}$, i.e., a conjunction of $e$ equations and $i$ inequalities.
A \emph{solution} of a NCSP is an assignment of its variables $v \in \vv_0$ that satisfies its constraints. The \emph{solution set} $\Sigma$ of a NCSP is the region within its initial domain that satisfies its constraints, i.e., $\Sigma(\vv_0) := \{\tilde{v}\in \vv_0 ~|~ c(\tilde{v})\}$.
The target of this paper is
\emph{under-constrained} NCSPs such that $n > e$.
In general, an under-constrained problem may have a continuous set of infinitely many solutions.

\Subsubsection{Branch and Prune Algorithm.}
\label{s:bap}

\begin{figure}[t]
\begin{minipage}{0.32\textwidth}
  \begin{center}
    \includegraphics[width=0.9\textwidth]{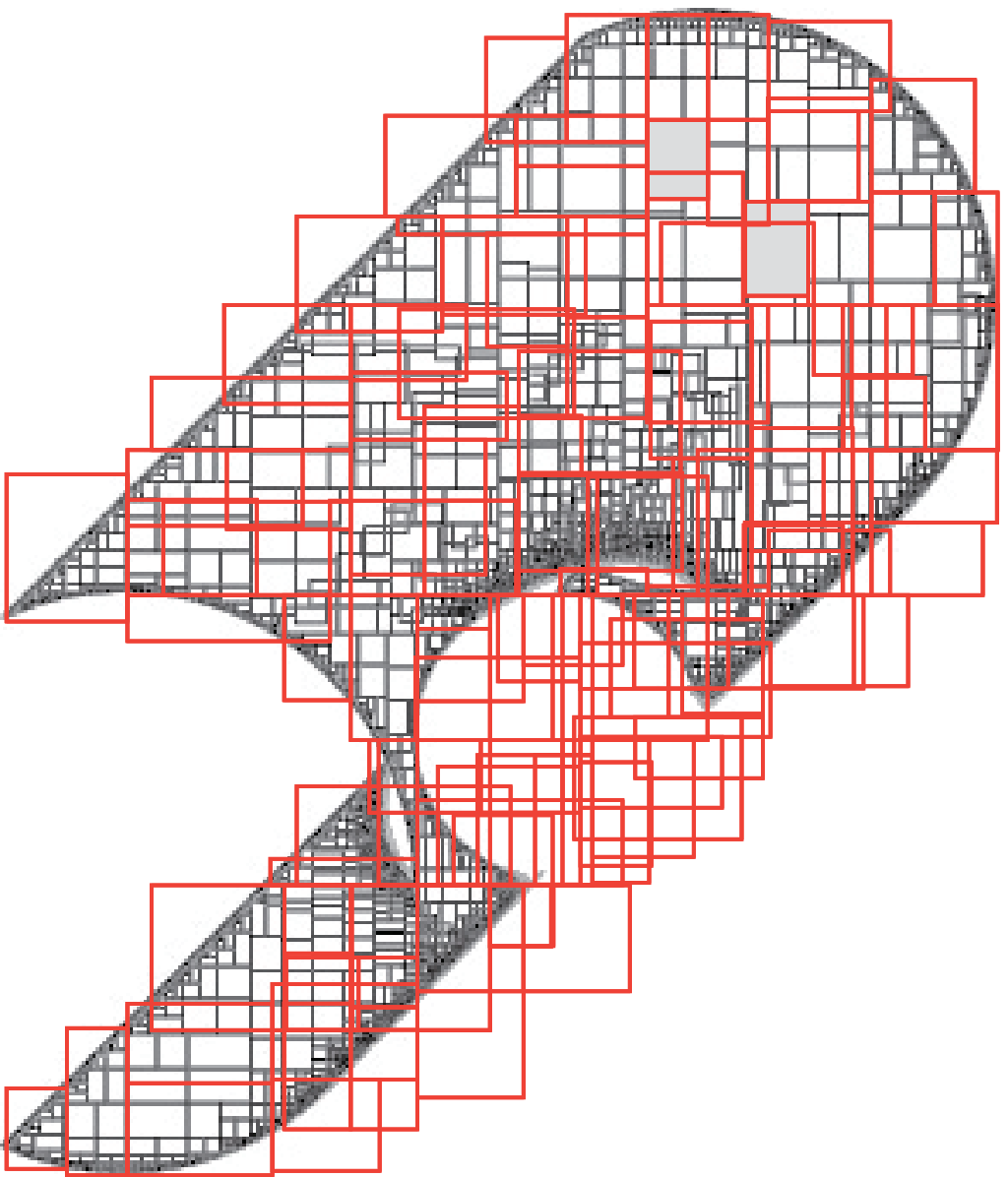}
    \vspace{-1em}
    \caption{\label{f:paving}Overlay of two solution box sets (pavings) with $\epsilon=0.01,0.1$}
  \end{center}
\end{minipage}
\hspace{.2em}
\begin{minipage}{0.28\textwidth}
  \begin{center}
    \vspace{1.5em}
    \includegraphics[width=0.8\textwidth]{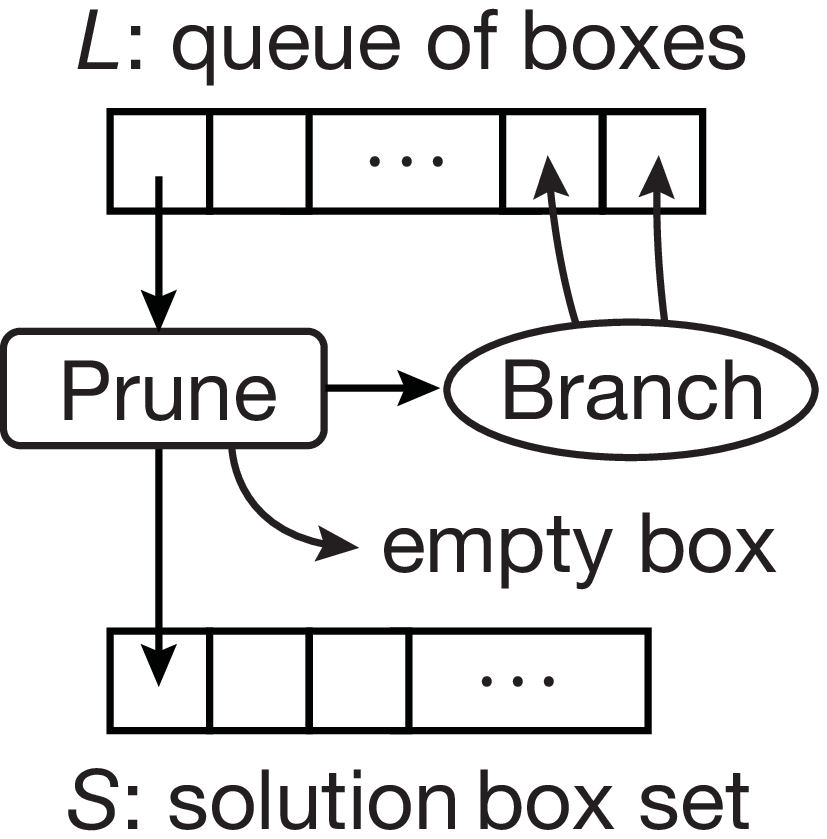}
    \vspace{.7em}
    \caption{\label{f:bap}A step computation of the branch and prune algorithm}
  \end{center}
\end{minipage}
\hspace{.2em}
\begin{minipage}{0.35\textwidth}
  \begin{center}
    \includegraphics[width=\textwidth]{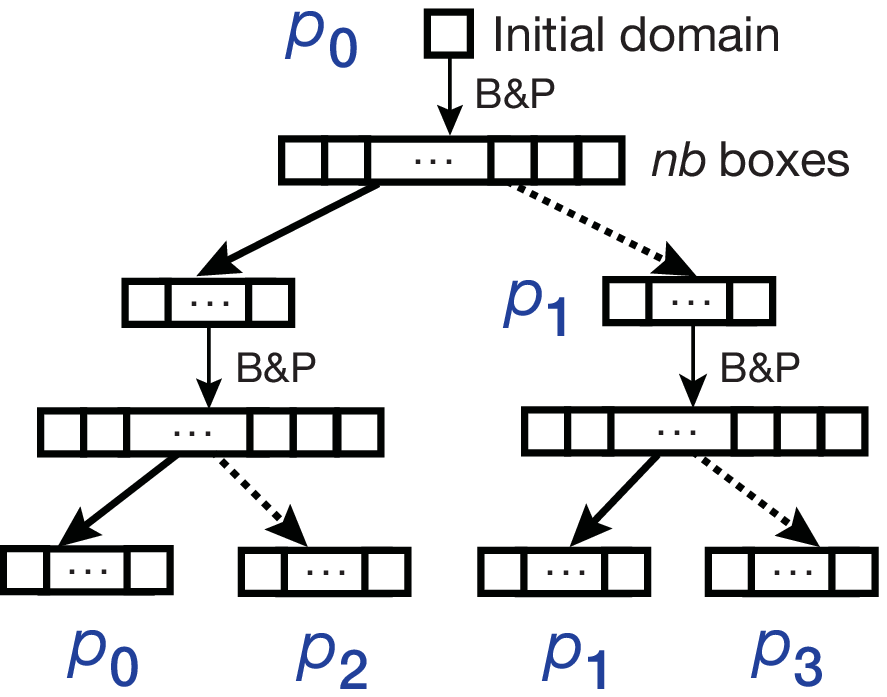}
    \vspace{-1.9em}
    \caption{\label{f:pp}Distribution of the box queue $L$ in the preprocess}
  \end{center}
\end{minipage}
\end{figure}

The \emph{branch and prune algorithm}~\cite{VanHentenryck1997} is the standard solving method for NCSPs. 
It takes a NCSP and a precision $\epsilon$ as an input and outputs a set of boxes (or \emph{paving}) $\BoxSet$ that approximates the solution set with precision $\epsilon$. 
Examples of $\BoxSet$ are illustrated in Figure~\ref{f:paving}.

An intermediate state of the algorithm is represented as a pair of sets of boxes $(L,S)$.
The solver receives an initial state $(\{\VVInit\},\emptyset)$
and iteratively applies the \emph{step computation} (illustrated in Figure~\ref{f:bap})
 until it reaches a final state $(\emptyset,\BoxSet)$.
%
In the step computation,
first, it takes the first element of the queue $L$ 
of boxes
and applies the $\Prune$ procedure, which is a filtering procedure that shaves boundary portions of the considered box.
In this work, we use an implementation proposed in \cite{Ishii2012} which provides a verification process based on an interval Newton method combined with a relatively simple filtering process based on the Hull consistency\cite{Benhamou1999hullbox}.
\Todo{As a result, a box becomes either empty, precise enough (its width is smaller than $\epsilon$), verified as an \emph{inner} box of the solution set $\Sigma$, or undecided.
Precise and inner boxes are appended to $S$ and undecided boxes are passed to $\Branch$.
}
Second, the $\Branch$ procedure bisects the box at the midpoint along a
component corresponding to one of the variables and the sub-boxes will be put back in the queue.
In this work, we assume $\Branch$ selects variables in an order which makes
the search to behave in a breadth-first manner and thus 
the solving process gradually improves the precision of the overall pavings (Figure~\ref{f:paving}).

The computation of
$\Prune$ is expensive and is the bottleneck of the solving process.
Under certain conditions, application of $\Prune$ contracts a large portion of the search space into a tight box (cf. quadratic convergence of the interval Newton methods).
$\Prune$ can also filter out the whole box if the considered box is verified as an inner or totally inconsistent region.
These characteristics of $\Prune$ result in the unbalanced nature of search trees.
\Todo{Therefore, a straightforward parallel method does not work efficiently.
It is crucial for efficient NCSP solving to execute $\Prune$ on each step of traversing the search tree which makes it more difficult to distribute a search path among processors.
These properties will be discussed in Section~\ref{s:discussions}.
}

\section{Parallel Branch and Prune}
\label{s:parallelization}

We propose a parallel method that runs several workers on the available CPU cores $p_0,\ldots, p_{\#p-1}$ (we assume a single worker runs on a core and identify both a worker and a core with $p_i$).
Workers homogeneously interleave the following three procedures and cooperate in a decentralized fashion:
(i) breadth-first branch and prune search, 
(ii) distribution and workload balancing of search space
in a sender-initiated manner, 
and (iii) termination detection.
%
%
Sub-trees in the search space of the branch and prune becomes unbalanced but can be searched independently: there are no confluence of multiple branches, and we have no shared information between branches.
Distribution of the search space among workers is done in preprocess and postprocess as described in the following subsections.
The preprocess distributes the portions of the search space to the other workers, 
and the postprocess balances the load of each worker during search.
Termination is detected by circulating a termination token via idling workers based on Dijkstra's method (see Section~11.4.4 of \cite{Grama2003}).
%

\Subsubsection{Preprocess: Search-Space Splitting and Distribution.}
\label{s:pre}

A solving process is started by a worker $p_0$ that possesses the initial domain (i.e., a box which contains the whole search space) in the queue $L$.
To distribute the subsequent search space (i.e., a queue of boxes) equally to each core, the preprocess invokes a partition of the queue in two (or more) and then sends a portion to another worker.
Figure~\ref{f:pp} illustrates in a downward direction some initial transitions of the box queue $L$ distributed among three workers.
%
In our implementation, the distribution routing is formed as a binary tree whose height is $\lceil \log_2 \# p \rceil$.
In each node of the tree, the branch and prune process runs until the number of boxes in the queue reaches $\NBB$. Next, the queue is sorted by the volume of the boxes and the half of the content (i.e., $\NBB / 2$ boxes) is sent to the other core via the right branch.


\Subsubsection{Postprocess: Dynamic Load Balancing.}
\label{s:post}

\Todo{
  During search, each worker normalizes the loads within a predefined \emph{neighborhood}
 which consists of a small number of \emph{neighbor} workers.
}
Because there are sufficiently large number of boxes, we simply regard the number of boxes in the queue $L$ as the amount of load.
Assume $\# p$ workers are running and each worker $p_i$ possesses $l_i$ boxes in its queue. 
We also assume for each worker $p_i$ that $N_i$ is a set of $|N|$ neighbor workers,
$N_i^{-1}$ is a set of workers
where $p_j\in N_i^{-1} \Leftrightarrow p_i \in N_j$,
$L_i$ is a set of loads of the neighbor workers,
and $\Delta$ is a predefined load margin.
The load balancing procedure of a worker $p_i$ performs the following steps once every $\NS$ branch and prune steps. 
\begin{enumerate}
\item For each worker $p_j$ in $N_i^{-1}$, inform the load $l_i$ and put in the list $L_j$.
\item Calculate the mean $\mu$ of the loads in $L_i$.
\item If $\mu < \Delta$, for each worker $p_j$ in $N_i$, send at most $\mu - l_j$ boxes to $p_j$ (to be efficient, a certain number of boxes should be kept locally).
\end{enumerate}
Neighbor workers can be identified e.g. as adjacent nodes in the $|N|$-dimensional mesh of workers.
\Todo{The routing between neighbor workers is fixed during a solving process and thus it may happen that a worker possesses an excessive load than others. However, this load imbalance will be resolved by the subsequent load balancing processes.}



\section{Experimental Results}
\label{s:exp}

We have implemented the proposed method and measured the speedup of the solving process of under-constrained NCSPs.
\Todo{The experiments were performed with an exhaustive set of parameter combinations to explore the optimal settings.}

\Subsubsection{Implementation.}

We have implemented the proposed method with C++ (gcc ver.~4.4.7 and 4.3.4) and X10
(ver.~2.3.1)\cite{X10:2005}, a high productivity language for parallel
computing.
In the following, we use the term \emph{place}, which is a notion of X10 that in our setting represents a CPU core.
%
Libraries Realpaver (ver.~1.1)\cite{LG2006} for sequential NCSP solving and Gaol (ver.~4.0.1)\footnote{\url{http://sourceforge.net/projects/gaol/}} for basic interval computation were used to facilitate the implementation.
The $\Prune{}$ procedure was realized by calling the sequential implementation in Realpaver.
Each run of $\Prune$ took around 0.2--1ms and the overall execution became the bottleneck of the branch and prune algorithm (occupied greater than $95\%$ of running time in sequential solving).
The procedures for search space distribution and load balancing were implemented with X10.
Communication of boxes and loads between places were implemented as \texttt{async} tasks and performed in parallel to the search process so that the overhead will be hidden.
\Todo{In the experiments, timings $t_1$ for sequential runs on single core were measured
using the C++ implementation described in \cite{Ishii2012},
which worked identically and faster than our X10 version.
}

\Subsubsection{Experiment Environments.}

Two sets of experiments were operated using (1) a shared-memory machine equipped
with 40 cores (four of 10-core Intel Xeon E7-4860 2.26GHz) and 256GB
of local memory and
(2) up to 256 cores of SGI UV1000, a pseudo-shared memory machine
equipped with 2,048 cores (8-core Xeon E7-8837 2.67GHz) and 16TB of memory.
UV1000 works as a single shared memory machine
by emulating memory accesses using communication based on
a high speed NUMAlink5 network which has a bandwidth of 120Gbps.
We used the MPI backend of X10 
with options $\texttt{X10\_NTHREADS}=6$ and $\texttt{GC\_NPROCS}=2$.

\Subsubsection{Experiments on a Shared Memory Machine.}

We solved the 
problems shown in \cite{Ishii2012,CCGIJ2014} using 40 cores of the machine (1).
\Todo{We report the results for two representative problems.}
Parameters in the load balancing method were set as either combination of the following values:
$\NBB=32, |N| \in \{2,4\}, \NS \in \{10,100,1000\}$, and $\Delta = 10$.
We also computed with and without the preprocess (when the preprocess is not used, the postprocess is executed from the beginning).
For each problem, we solved two instances with two multiplicative precisions.
The specification of each instance and the computational results are presented in Table~\ref{t:results}.
In the table, the columns ``problem'', ``size'', ``$\epsilon$'', ``pp'', 
and ``$|N|$'' represent the name of the problem, size (i.e., the number of projection/parameter variables), the precision, usage of the preprocess, 
and the number of neighbor workers, respectively.
The rest of the columns represents the results. 
$t_1$ and $\NBr{1}$ represent the running time and the number of branches on single core.
$t_{\NS,i}$ and $\NBr{\NS,i}$ represent the running time and the largest number of branches performed by a worker when computed with the interval $\NS$ and $i$ X10 places \Todo{(best timings are underlined)}. 
%
Figure~\ref{f:su:40} illustrates the speedup of the solving process.


\begin{table*}[t]
\begin{center}
  \caption{\label{t:results} Experimental result on the shared memory machine} 
	\vspace{-.5em}
    \begin{tabular}{l|c|r|c|c|r|r|r|r|r|r} \hline \hline
      problem & size & $\epsilon$ & 
	  pp & $|N|$ &
      $t_1$ & $t_{10,40}$ & $t_{100,40}$ & $t_{1000,40}$ & 
	  $\NBr{1}$ & $\frac{\NBr{1}}{\NBr{100,40}}$ \\
	  \hline
	  4D sphere & 2+4 & 0.02 & 
	  yes & 2 &
	  254 & 9.52 & \underline{7.94} & 10.1 &
	  238\,319 & 37.0 
	  \\ 
	  and plane & & & 
	  & 4 &
	  & 11.0 & 8.73 & 9.45 & 
	  & 37.6 
	  \\ 
	  (\textit{sp2-4}) & & & 
	  no & 2 &
	  & 9.34 & 8.13 & 16.5 &
	  & 35.4 
	  \\ 
	  & & & 
	  & 4 &
	  & 10.8 & 8.67 & 10.5 & 
	  & 37.3 
	  \\ 
	  \cline{3-11}
	  & & 0.01 & 
	  yes & 2 &
	  721 & 38.6 & 22.8 & 23.9 & 
	  669\,601 & 38.0 
	  \\ 
	  & & & 
       & 4 &
	  & 38.1 & 25.1 & 24.0 & 
	  & 38.7 
	  \\ 
	  & &  & 
	  no & 2 &
	  & 35.7 & \underline{22.7} & 30.0 & 
	  & 37.6 
	  \\ 
	  & & & 
       & 4 &
	  & 36.8 & 24.7 & 24.7 & 
	  & 38.7 
	  \\ 
	  \hline
	  3-RPR robot & 3+3 & 0.2 & 
	  yes & 2 &
	  1\,100 & 199 & 73.9 & 34.1 &
	  1\,936\,939 & 33.7 
	  \\ 
	  (\textit{3rpr}) & & & 
	  & 4 &
	  & 296 & 68.5 & 36.4 & 
	  & 38.0 
	  \\ 
	   & & & 
	  no & 2 &
	  & 185 & 64.0 & \underline{34.0} & 
	  & 33.5 
	  \\ 
	  & & & 
	  & 4 &
	  & 244 & 58.3 & 36.1 & 
	  & 38.0 
	  \\ 
	  \cline{3-11}
	  & & 0.1 & 
	  yes & 2 &
	  4\,080 & 1\,010 & 714 & 282 & 
	  7\,186\,845 & 30.2 
	  \\ 
	  & & & 
	  & 4 &
	  & 2\,820 & 1\,070 & 257 & 
	  & 36.0 
	  \\ 
	  & & & 
	  no & 2 &
	  & 971 & 678 & 244 & 
	  & 28.5 
	  \\ 
	  & & & 
	  & 4 &
	  & 2\,630 & 901 & \underline{231} & 
	  & 36.0 
	  \\ 
	  \hline
    \end{tabular}
\end{center}
\end{table*}

\Subsubsection{Experiments on a Cluster with High-speed Interconnection.}

\begin{figure}[t]
\begin{center}
  \includegraphics[height=0.065\textheight]{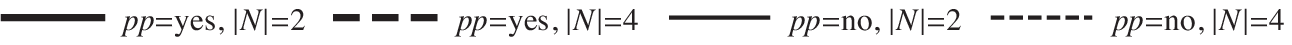}

  \vspace{-3em}
  \subfigure[$\textit{3rpr}, \epsilon=0.2$.]
  {\includegraphics[width=0.45\textwidth]{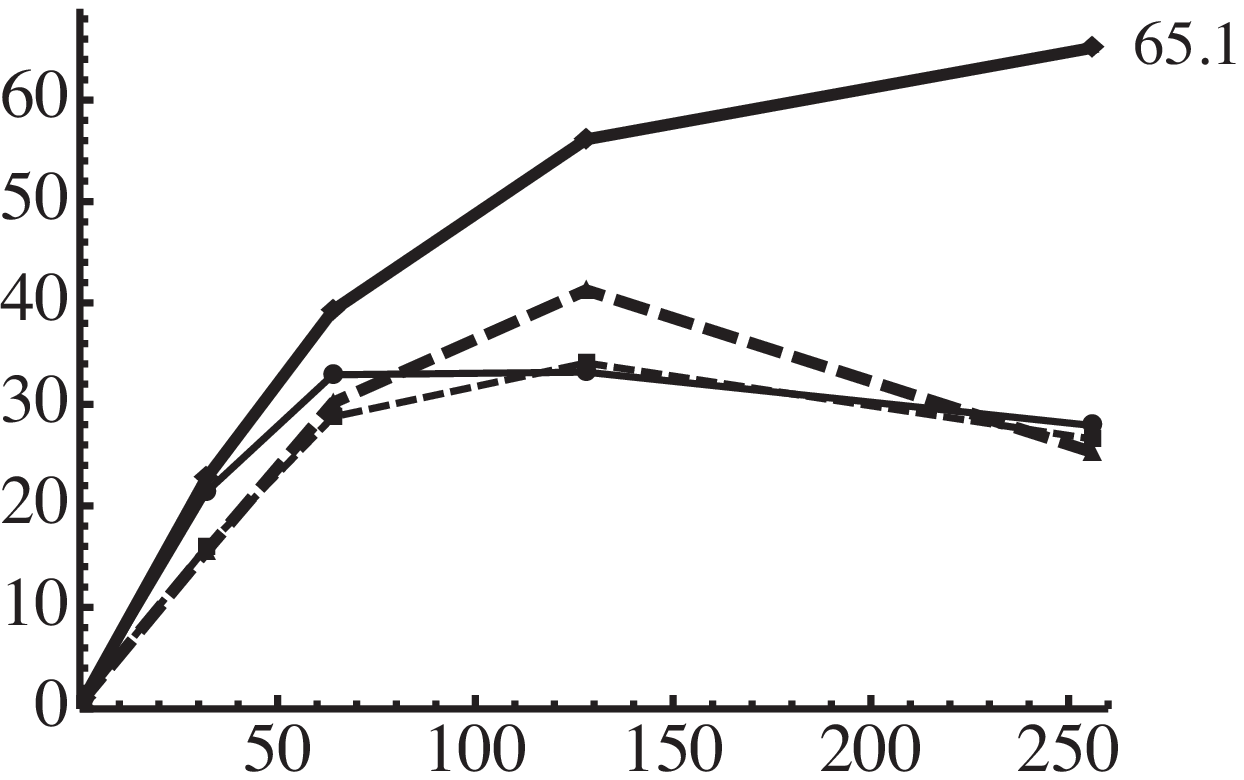}
    \label{f:uv1000:3rpr:02}}
  \hspace{.5em}
  \subfigure[$\mathit{3rpr}, \epsilon=0.1$.]
  {\includegraphics[width=0.45\textwidth]{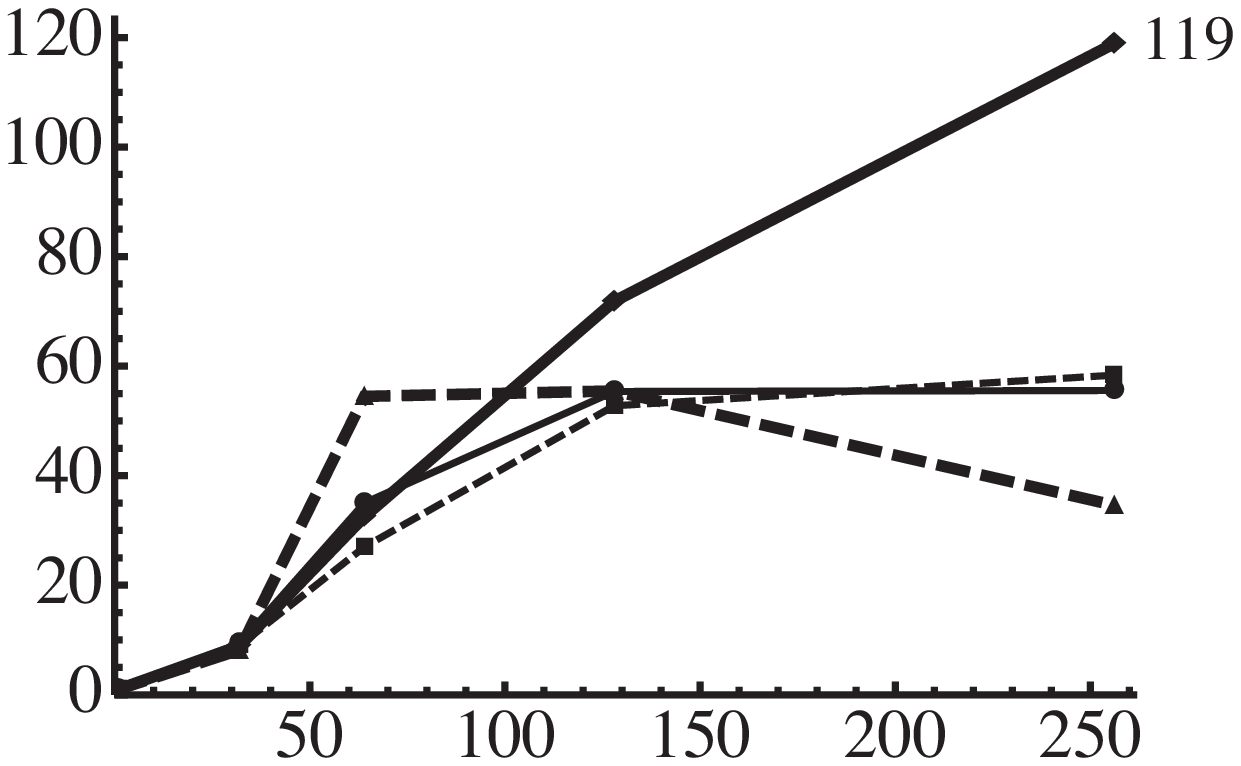}
    \label{f:uv1000:3rpr:01}}
  \caption{\label{f:su:uv1000} Speedups using 256 cores of UV1000 
  }
\end{center}
\end{figure}

We solved the problem ``$\mathit{3rpr}$'' using up to 256 cores of the machine (2), UV1000.
Parameters in the load balancing method were set as either combination of the following values:
$\NBB=8, |N| \in \{2,4\}$, $\NS=1000$, and $\Delta = 10$.
The results are presented in Table~\ref{t:results:uv1000}.
Each column of the table represents the same information as presented in Table~\ref{t:results} except that 
the column ``$\NComm{1000,256}$'' represents the number of loads sent by the load balancer in the solving process with $\NS=1000$ and 256 X10 places.
Figure~\ref{f:su:uv1000} illustrates the speedup of the solving process.

\begin{table*}[t]
\begin{center}
  \caption{\label{t:results:uv1000} Experimental result on the UV1000 cluster}
	\vspace{-.5em}
    \begin{tabular}{l|c|r|c|c|r|r|r|r|r} \hline \hline
      problem & size & $\epsilon$ & 
	  pp & $|N|$ &
      $t_1$ & $t_{1000,32}$ & $t_{1000,256}$ & 
	  $\frac{\NBr{1}}{\NBr{1000,256}}$ & 
	  $\NComm{1000,256}$ \\ 
	  \hline
      3-RPR robot & 3+3 & 0.2 & yes & 2 &
	  850 & 37.4 & \underline{14.0} & 
	  131 & 
	  18\,648 \\ 
	  ($\mathit{3rpr}$) & & & & 4 &
	  & 54.6 & 33.6 & 
	  157 & 
	  85\,904 \\ 
	  & & & no & 2 &
	  & 39.8 & 20.4 & 
	  43.8 & 
	  10\,856 \\ 
	  & & & & 4 &
	  & 53.4 & 32.0 & 
	  123 & 
	  66\,212 \\ 
 \cline{3-10}
	  & & 0.1 & yes & 2 &
	  3\,040 & 341 & \underline{25.6} & 
	  192 & 
      39\,608 \\ 
	  & & & & 4 &
	  & 371 & 87.8 & 
	  176 & 
      128\,084 \\ 
	  & & & no & 2 &
	  & 325 & 54.7 & 
	  105 & 
      34\,892 \\ 
	  & & & & 4 &
	  & 339 & 52.1 & 
	  184 & 
      129\,512 \\ 
	  \hline
    \end{tabular}
\end{center}
\end{table*}

\subsection{Discussions}
\label{s:discussions}

In the experiments, our method scaled up to 256 cores with the optimal configurations.
We achieved speedups up to 32.3 fold using 40 cores of the shared memory machine and up to 119 fold using 256 cores of the cluster machine.

The best speedup of 119 fold was obtained with the preprocess.
The preprocess facilitates and accelerates the workload distribution in the early stage of the search process.
In some of the experiments without using the preprocess, the speedup ratio became saturated when using many cores (e.g., $\textit{sp2-4}$ with $\NS=1000$ and the experiments on the cluster).
This was because the load balancing process was too infrequent for the given number of workers and the work load diffusion became too slow.
When comparing the right-hand graphs for the instance $\textit{sp2-4}, \NS=1000$, in Figure~\ref{f:su:40}, we can notice that the point of saturation shifts according to the search space size.
On the other hand, in some other experiments, the results got worse with the preprocess (e.g., results with $\NS=10$ on the machine (1)).
It occasionally happens that the preprocess mostly solves the problem.
However, the preprocess can result in highly unbalanced search trees
because of the $\Prune{}$ process, and
in such cases the postprocess will not have enough time for load balancing.

\afterpage{%
\begin{figure}[H]
\begin{center}
  \includegraphics[width=0.9\textwidth]{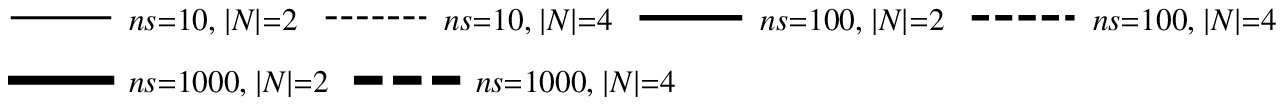}
  \subfigure[$\textit{sp2-4}, \epsilon=0.02$.]
  { \begin{minipage}{\textwidth}
    \begin{center}
\includegraphics[width=0.43\textwidth]{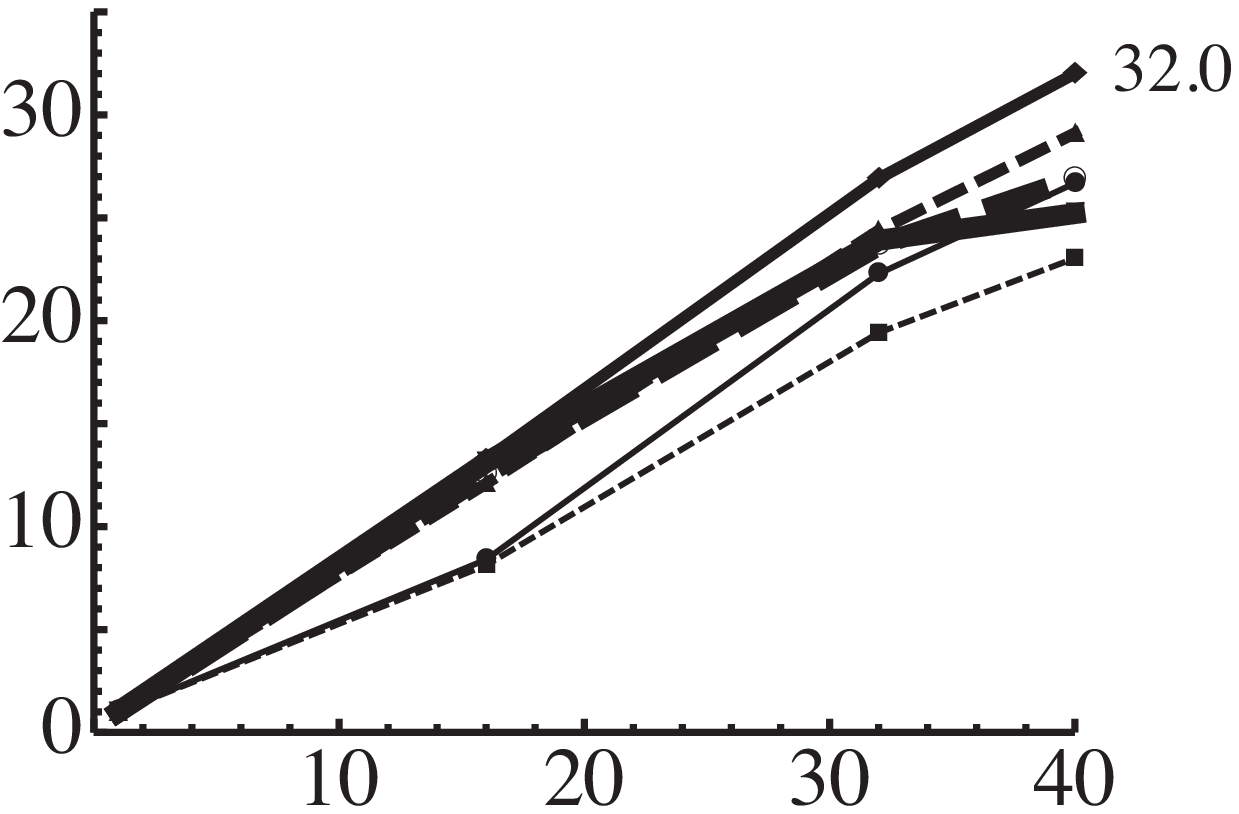}
\hspace{1em}
\includegraphics[width=0.43\textwidth]{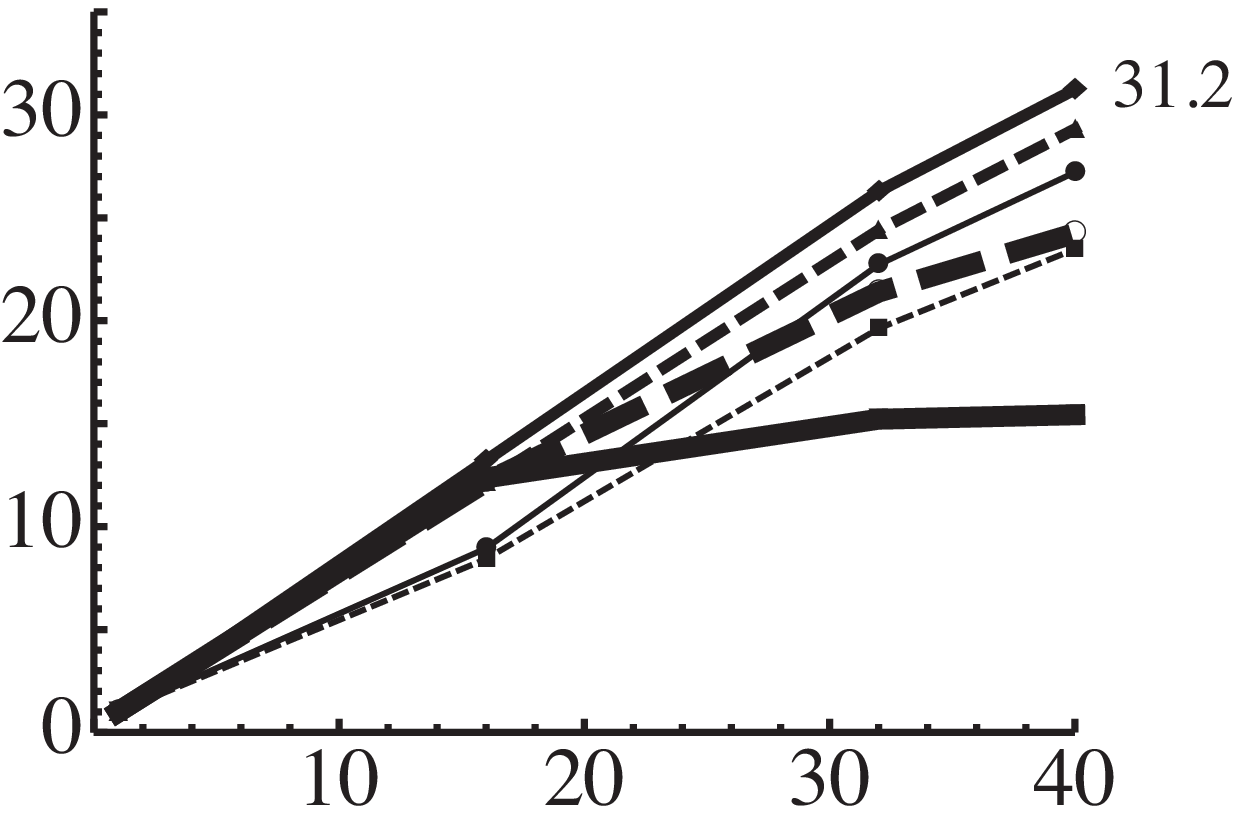}
    \end{center}
    \end{minipage}}
  \subfigure[$\textit{sp2-4}, \epsilon=0.01$.]
  { \begin{minipage}{\textwidth}
    \begin{center}
\includegraphics[width=0.43\textwidth]{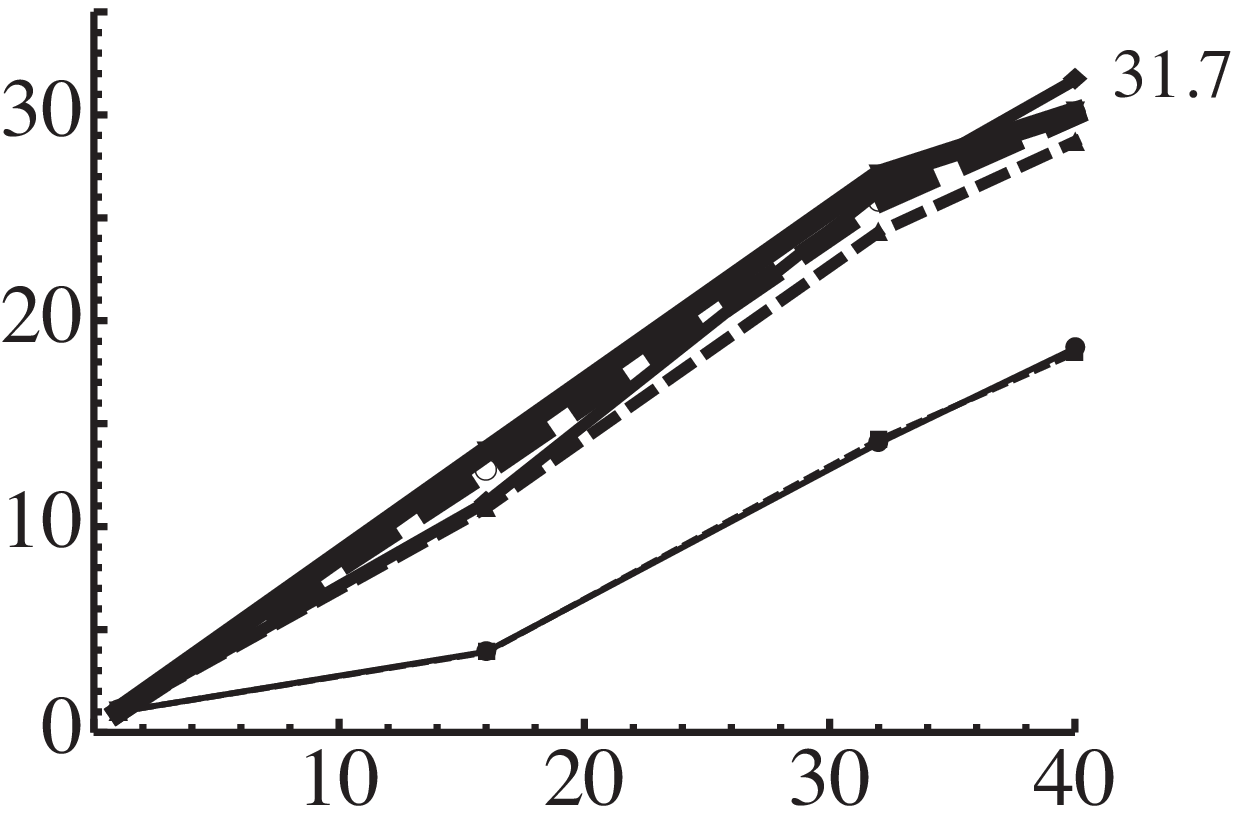}
\hspace{1em}
\includegraphics[width=0.43\textwidth]{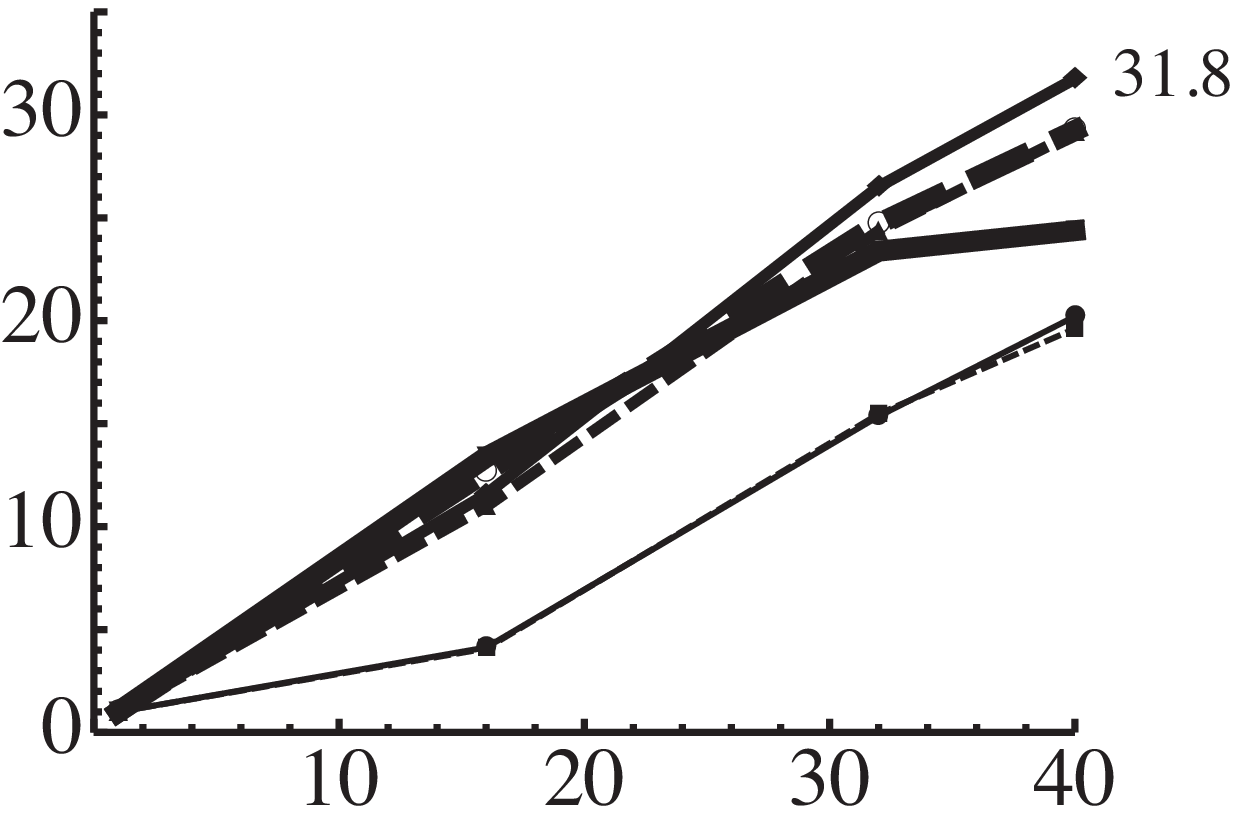}
    \end{center}
    \end{minipage}}
  \subfigure[$\textit{3rpr}, \epsilon=0.2$.]
  { \begin{minipage}{\textwidth}
    \begin{center}
\includegraphics[width=0.43\textwidth]{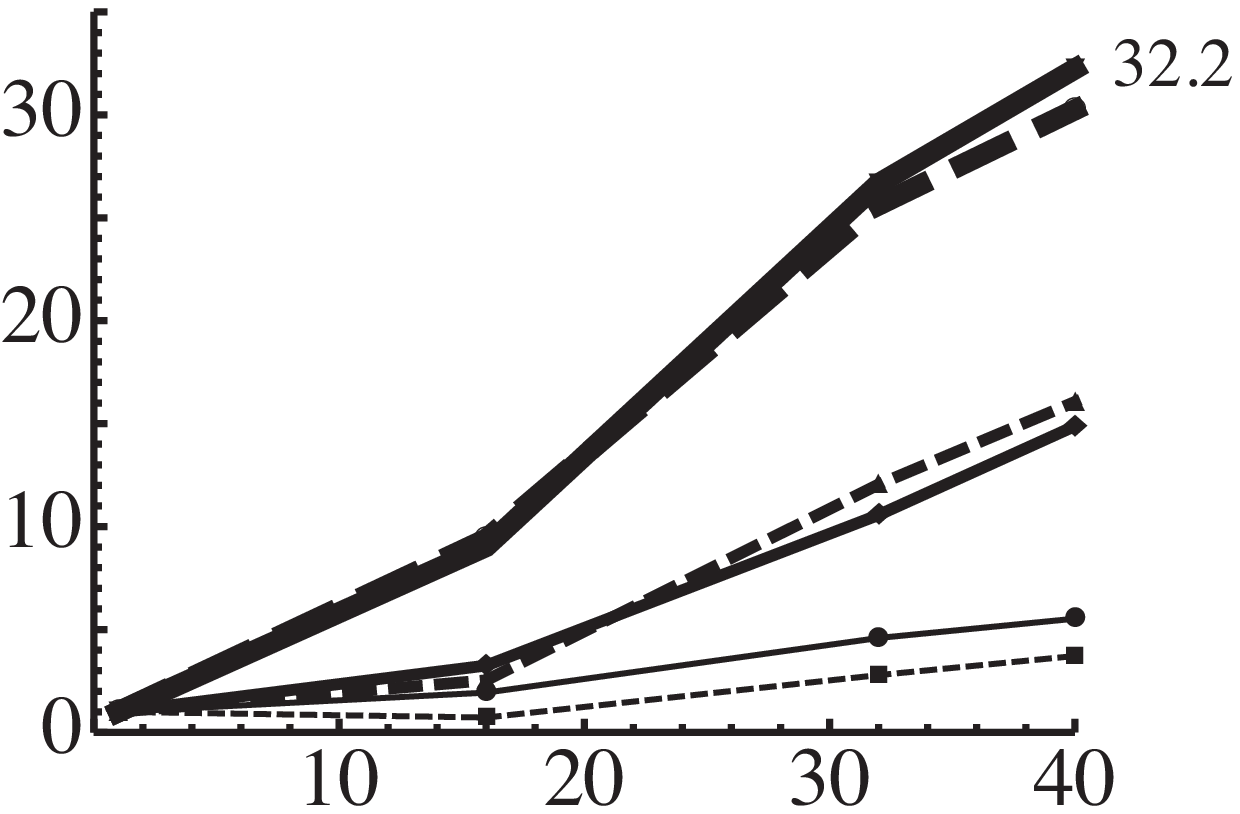}
\hspace{1em}
\includegraphics[width=0.43\textwidth]{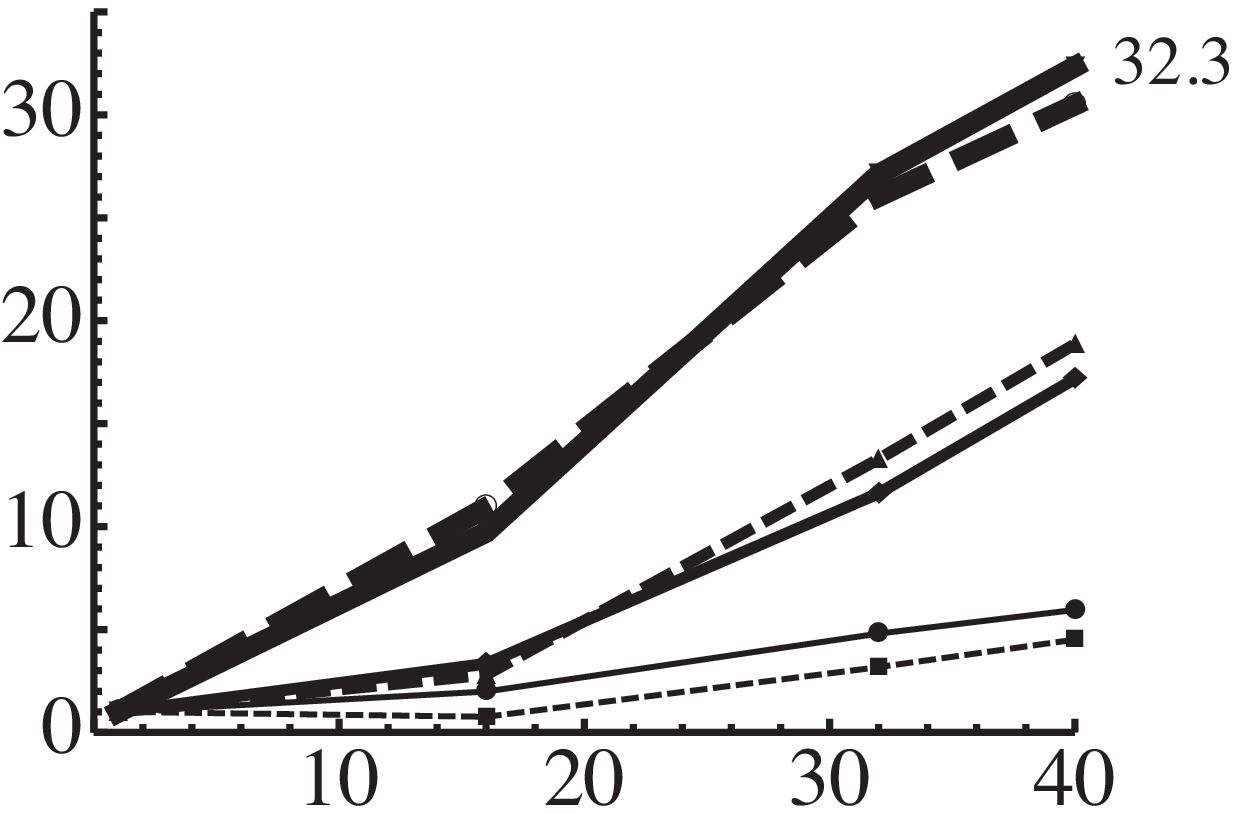}
    \end{center}
    \end{minipage}}
	\subfigure[$\textit{3rpr}, \epsilon=0.1$. \label{f:su:40:3rpr}]
  { \begin{minipage}{\textwidth}
    \begin{center}
\includegraphics[width=0.43\textwidth]{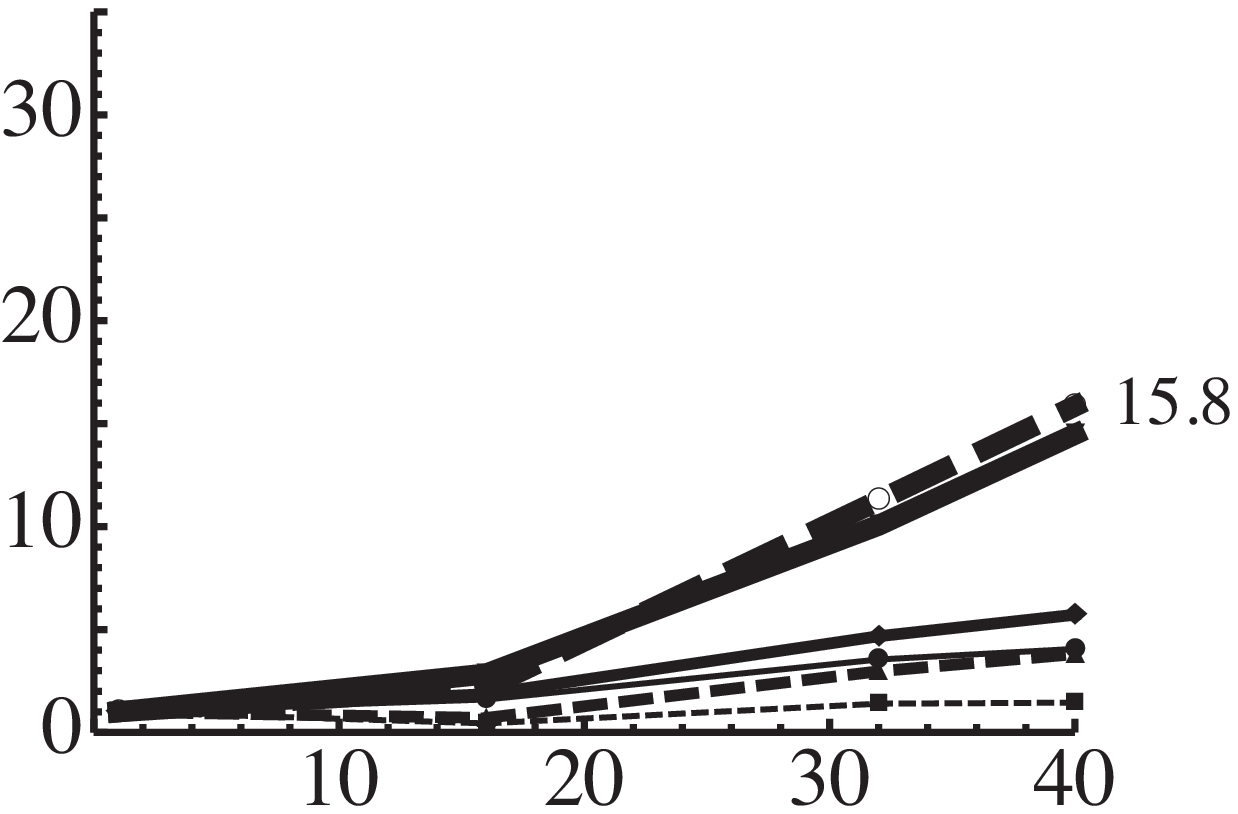}
\hspace{1em}
\includegraphics[width=0.43\textwidth]{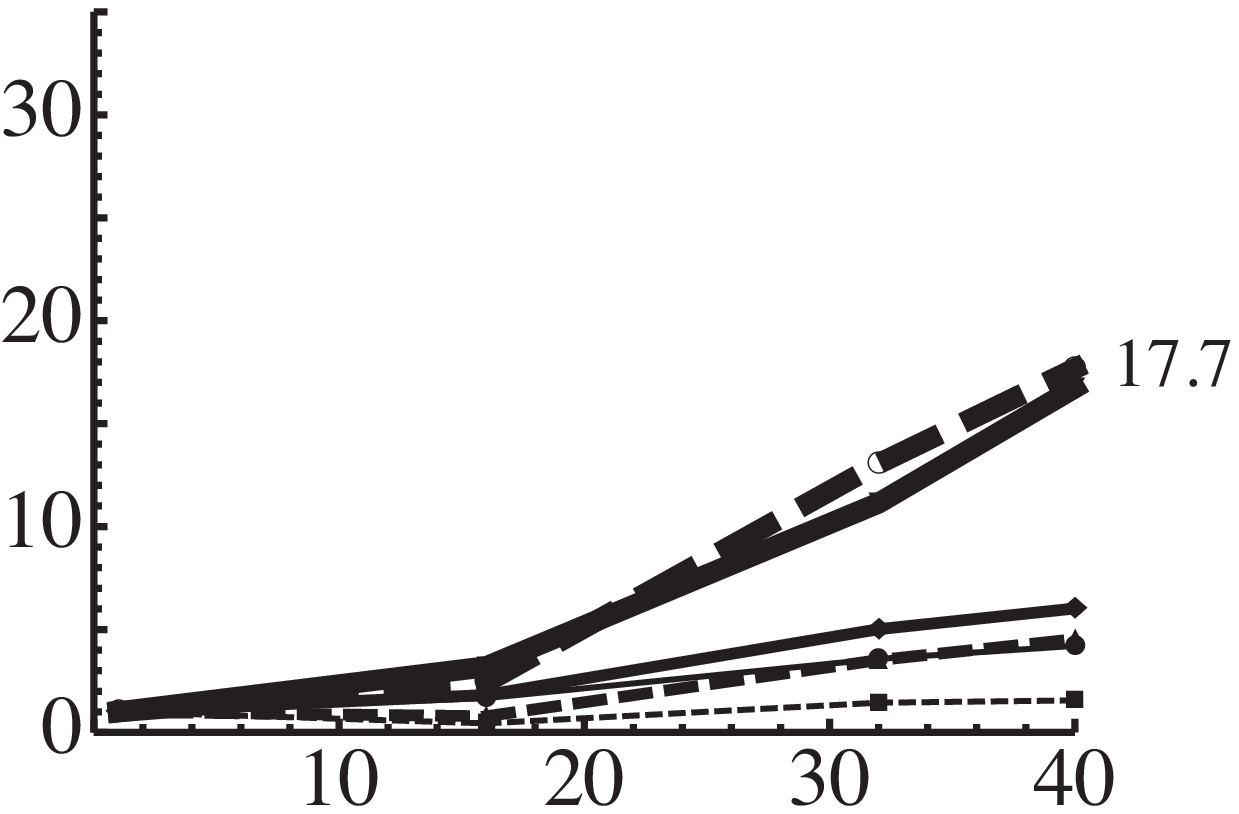}
    \end{center}
    \end{minipage}}
  \caption{\label{f:su:40} Speedups on the shared memory machine with 40 cores. 
	  Left- and right-hand side graphs correspond to computations with and without the preprocess, respectively
  }
\end{center}
\end{figure}
}

Regarding the neighborhood sizes $|N|=2,4$, there was a trade off between the workloads balance and the amount of communications required.
For the shared memory machine, it was unclear which size had the advantage.
However, for UV1000, the solver was notably slower for $|N|=4$ than $|N|=2$.
It is understandable because larger number of neighbors significantly increased
the number of communications (see ``$\NComm{1000,256}$'' in Table~\ref{t:results:uv1000}) and
communications between places were much more costly compared to normal shared
memory machines despite the high speed network of UV1000.

Three intervals $\NS=10,100,1000$ were used for load balancing which determined the
speed of workload distribution.
When the distribution was too slow, the speedup ratio did not scale well
(e.g., $\textit{3rpr}$, $\epsilon=0.2$, with $\NS=1000$ on the machine (1)).
Conversely, small intervals required greater amount of communications and
therefore we used $\NS=1000$ to draw better performance on the cluster 
where communications were more costly.

\Todo{
There was a large overhead caused by the workers sending a large number of boxes
for load balancing when the number of workers was not sufficient against the
problem size.
Speedups for \textit{3rpr}, $\epsilon=0.1$, using 40 workers or less
shows an example of such overheads (Figure~\ref{f:su:40:3rpr}).
Resolving this overhead by suppressing redundant box sends is a part of the
future work.
}

\section{Conclusions}
\label{s:concl}

In this paper, we proposed a parallel branch and prune algorithm,
\Todo{based on, non-portfolio, search-space splitting approach}.
In the experiments, using 256 X10 places (i.e., cores), we achieved speedup factors of as much as 119.
%
We expect that our parallelized solver will be applied to large practical problems, e.g., the robotics problems in \cite{CCGIJ2014}.

\paragraph{Acknowledgments.}
This work was partially funded by JSPS (KAKENHI 25880008 and \Todo{25700038}).
Computing resources for the experiments in this paper were provided by
Prof. Kazunori Ueda (Waseda University, Tokyo) and 
a Compute Canada RAC award, for environments (1) and (2), respectively.

\bibliographystyle{splncs03}
\bibliography{parallel}


\end{document}